\documentclass{article}

\PassOptionsToPackage{square,numbers}{natbib}
\usepackage[final]{neurips_2021}
\usepackage[utf8]{inputenc} 
\usepackage[T1]{fontenc}    
\usepackage{hyperref}       
\usepackage{url}            
\usepackage{booktabs}       
\usepackage{amsfonts}       
\usepackage{nicefrac}       
\usepackage{microtype}      
\usepackage{xcolor}         

\usepackage{times}
\usepackage{doi}
\usepackage{graphicx}
\usepackage{bm}
\usepackage{amsmath, amssymb,dsfont}
\usepackage{comment}

\def\eqref#1{equation~\ref{#1}}

\def\1{\bm{1}}

\def\vd{{\bm{d}}}

\def\vx{{\bm{x}}}

\def\vz{{\bm{z}}}

\def\evd{{d}}

\def\evx{{x}}

\def\mE{{\bm{E}}}

\def\mH{{\bm{H}}}

\DeclareMathAlphabet{\mathsfit}{\encodingdefault}{\sfdefault}{m}{sl}
\SetMathAlphabet{\mathsfit}{bold}{\encodingdefault}{\sfdefault}{bx}{n}

\def\emE{{E}}

\newcommand{\E}{\mathbb{E}}

\newcommand{\R}{\mathbb{R}}

\DeclareMathOperator{\kl}{KL}
\DeclareMathOperator{\lstm}{\textsc{Lstm}}

\title{Preventing posterior collapse \\ in variational autoencoders for text generation  \\ via decoder regularization}

\author{%
Alban Petit \\
Université  Paris-Saclay, CNRS, LISN, 91400, Orsay, France \\
\texttt{alban.petit@lisn.upsaclay.fr}
\AND
Caio Corro \\
Université  Paris-Saclay, CNRS, LISN, 91400, Orsay, France \\
\texttt{caio.corro@lisn.upsaclay.fr}
}

\begin{document}

\maketitle

\begin{abstract}
  Variational autoencoders trained to minimize the reconstruction error are sensitive to the posterior collapse problem, that is the proposal posterior distribution is always equal to the prior.
  We propose a novel regularization method based on fraternal dropout to prevent posterior collapse.
  We evaluate our approach using several metrics and observe improvements in all the tested configurations.
\end{abstract}

\section{Introduction}

Deep generative models like Variational Autoencoders (VAE) \cite{kingma2013} and Generative Adversarial networks (GAN) \cite{goodfellow2014gan}, among others, enjoy great popularity in many applications of machine learning, including natural language processing (NLP).
Unlike GANs, VAEs can manipulate discrete observed variables which makes them suitable for applications in text generation.
In these models, a sentence is generated according to the following process:
\begin{equation*}
    \vz \sim p(\vz), \quad \evx_1 \sim p_\theta(\evx_1 | \vz), \quad \evx_2 \sim p_\theta(\evx_2 | \vz, \evx_1) \quad \evx_3 \sim p_\theta(\evx_3 | \vz, \evx_1, \evx_2), \quad \dots 
\end{equation*}
where $\vz \in \R^k$ is a latent sentence representation (or sentence embedding) and $\evx_1, \evx_2, \dots \in X$ are observed words (or the generated sentence), with $X$ the vocabulary.
The generation stops when a special end of the sentence word is generated.
Without loss of generality, we assume that the prior distribution $p(\vz)$ is fixed.
The subscript $\theta$ indicates the parameters of the conditional distribution and, in our case, $\theta$ corresponds to the parameters of a neural network.
It is important to note that we do not make any independence assumption in the conditional distribution $p(\evx_t | \vz, \vx_{<t})$.

Training aims to search for parameters $\theta$ that maximize the likelihood of observed data, also called the evidence:
\begin{equation}
    \max_{\theta}~\E_{\tilde p(\vx)} \left[ \log p_{\theta}(\vx) \right]
    = \max_{\theta}~\E_{\tilde p(\vx)} \left[ \int \log p_{\theta}(\vx | \vz) p(\vz) d\vz \right]
    = \max_{\theta}~\E_{\tilde p(\vx)} \left[ \mathcal L(\vx, \theta) \right]
    \label{loglike}
\end{equation}
where $\tilde p$ is the empirical training data distribution.
In general, the objective function $\mathcal L$ is intractable because of the marginalization over latent variables $\vz$.
Variational methods propose to introduce a proposal distribution $q_\phi(\vz | \vx)$ to create a surrogate objective called the evidence lower bound (ELBO) defined as follows:
\begin{equation}
    \mathcal{E}(\vx, \theta, \phi) = \underbrace{\E_{q_\phi(\vz | \vx)}[\log p_\theta(\vx | \vz)]}_{\text{reconstruction term}} - \underbrace{\kl[q_\phi(\cdot | \vx) | p(\cdot)]}_{\text{divergence with prior}}
    \label{elbo}
\end{equation}
where the $\mathcal{E}$ defines a family of lower bounds on $\mathcal L$ parameterized by $\phi$, i..e\ $\forall \phi: \mathcal{E}(\vx, \theta, \phi) \leq \mathcal L(\vx, \theta)$.
During training, it is important to search for proposal parameters $\phi$ that gives the best bounds (i.e. maximizes the bounds), hence the name variational.
The new training problem is then:
\begin{equation}
    \max_{\theta, \phi}~\E_{\vx \sim \tilde p(\vx)} \left[ \mathcal{E}(\vx, \theta, \phi) \right]
    \label{objectif}
\end{equation}
The Expectation-Maximization (EM) algorithm \cite{dempster1977em} solves this problem via block coordinate ascent, i.e.\ maximizing successively with respect to $\phi$ (E step) and $\theta$ (M step).
Unlike EM, the VAE approach consists in optimizing this problem by joint stochastic gradient ascent over $\phi$ and $\theta$.
Moreover, unlike in the standard applications of EM, neither the distribution nor the family of the posterior $p_\theta(\vz | \vx)$ are known.
As a result, an independence assumption is made over the coordinates of $\vz$ in the distribution $q_\phi$ (also called mean field distribution).
Finally, the distribution $q_\phi$ is amortized over the data and parameterized by a neural network.
It is worth noting that during training, the reconstruction term in equation~\ref{elbo} is estimated via Monte-Carlo method using a single sample.
It is usual to call the distribution $q_\phi(\vz | \vx)$ that generates a latent representation from a sentence the \textbf{encoder} and the distribution $p_\theta(\vx | \vz)$  that reconstructs the original sentence the \textbf{decoder}.

Unfortunately in practice, VAEs for automatic text generation are sensitive to posterior collapse \cite{bowman2016}.
Informally, this means that the proposal distribution is not optimized correctly and remains close to the prior distribution for all data points: $\forall \vx: q_\phi(\vz | \vx) \simeq p(\vz)$.
This leads to a poor approximation of the objective~\ref{loglike} by the ELBO.
In the end, the decoder ignore that latent variable and no sentence representation $\vz$ is learned.

Previous work on the posterior collapse problem can be classified in two categories:
\begin{enumerate}
    \item On the one hand, modifications to the objective function were suggested.
    \citet{bowman2016} reweight the divergence with the prior term to temper its importance during training.
    \citet{kingma2016} and \citet{pelsmaeker2019vae} proposed to introduce constraints to force the divergence term to be greater than a hyperparameter.
    Finally, alternative surrogate objectives to train VAEs in the context of text generation have been proposed \cite{livne2020sentencemim,havrylov2020vae}.
    \item On the other hand, several authors proposed to modify the architecture of the decoder so that they are forced to rely on latent variable information.
    \citet{yang2017cnnvae} replace the recurrent neural network in the decoder by a convolutional neural network.
    \citet{dieng2019skipvae} proposed to use skip connections between the latent representation and the various hidden layers of the decoder.
\end{enumerate}

In this work, we propose a different approach to prevent the posterior collapse problem based on parameter regularization, that is without changing the generative model objective function or the architecture of the decoder.
Our contributions can be summarized as follows:
\begin{itemize}
    \item We propose to regularize the decoder parameters to prevent the posterior collapse. In particular, we use fraternal dropout \cite{zolna2018fraternal} to force the decoder to use the latent representation.
    \item We experiment our approach in various settings and report improved results with respect to several evaluation metrics.
\end{itemize}
We hope to encourage future research to explore this direction to improve VAEs for text generation.


\section{Decoder regularization: fraternal dropout}

We first give an intuitive motivation for our approach.
LSTMs have been widely used as neural language models and achieve competitive results \cite{merity2017lm}.\footnote{Our method is agnostic of the decoder architecture. As such, the same methodology can be directly applied to self-attentive networks/transformers \cite{vaswani2017transformers}.}
Language models are trained via the same reconstruction term as the one used in VAEs, showing that these neural architectures can efficiently maximize this term in the absence of latent variables, i.e.\ it can be maximized while ignoring the latent variables values.
To bypass this issue, we propose to introduce a regularization term in the objective that forces the hidden representations computed by the LSTM to be similar even if different words in the input are masked.
The decoder is then forced to rely on latent variable information.
To this end, we propose to rely on fraternal dropout \cite{zolna2018fraternal}.

The reconstruction term of the ELBO maximizes the log-likelihood using the usual teacher forcing technique for language models:
during training, the auto-regressive model is trained to predict the next word based on gold previous words.
Let $\vx \in X^n$ a sentence of length $n$.
Each word $\evx_i$ is represented by a vector taken from an embedding table.
We denote all of these vectors by a matrix $\mE \in R^{w \times n}$ where $w$ is the word embedding dimension.
A contextual representation is computed for each position in the sentence using a LSTM:
$$
    \mH = \lstm(\mE, \vz; \theta)
$$
where $\mH \in \R^{d \times n}$ is a matrix containing hidden representations of dimension $d$ for each sentence position.
The latent variable $\vz$ is projected and then given as an initialization for both the memory and the hidden state and also concatenated to each input.
We refer the reader to \cite{li2019} for more details on the architecture and the various parameters.

Word dropout \cite{dozat2017biaffine} consists in randomly replacing embeddings by a vector of zeros during training to prevent overfitting:
\begin{equation*}
    \vd \sim \mathcal{B}(b), \qquad \mE' \in \R^{w \times n} \text{~~t.q.~~} \emE'_{i, j} = \emE_{i, j} \evd_j, \qquad \mH = \lstm(\mE', \vz ; \theta)
\end{equation*}
where $\vd \in \{0, 1\}^n$ is a vector of booleans where each element is independently drawn from a Bernoulli distribution of parameter $b \in [0,1]$.
The matrix $\mE'$ corresponds to the matrix $\mE$ where a number of columns are filled with zeros.
Fraternal dropout consists in creating two matrices $\mE'$ and $\mE''$ as follows:
\begin{align*}
    &\vd \sim \mathcal{B}(b),\\
    &\mE' \in \R^{d \times n} \text{~~t.q.~~} \emE'_{i, j} = \emE_{i, j} \evd_j,
    &&\mE'' \in \R^{d \times n} \text{~~t.q.~~} \emE''_{i, j} = \emE_{i, j} (1 - \evd_j), \\
    &\mH' = \lstm(\mE', \vz ; \theta),
    &&\mH'' = \lstm(\mE'', \vz ; \theta).
\end{align*}
Matrices $\mE'$ and $\mE''$ are both used to compute the log-likelihood of the sentence and the mean log-likelihood then replaces the original reconstruction term.
Finally, a regularization term is introduced in the objective function:
\begin{equation*}
    \mathcal R(\theta ; \alpha) = - \alpha \| \lstm_\theta(\mE', \vz ; \theta) - \lstm_\theta(\mE'', \vz ; \theta) \|_2^2 
\end{equation*}
where $\alpha > 0$ is a hyperparameter.
Note that the regularization term forces the hidden representations computed by the LSTM to be similar with different masked inputs $\mE'$ and $\mE''$ but similar latent variables $\vz$.
Hence, the regularization term forces the decode to rely on latent variables $\vz$ to compute the LSTM's hidden representations.

\section{Experiments}

We evaluate our approach using the code distributed by by \citet{li2019}.
We kept the same hyperparameters as the ones used by the authors to avoid skewing the results in our favor.

We experiment with two datasets: Yelp \cite{shen2017} and the Stanford Natural Language Inference (SNLI) \cite{snli} data.
These two datasets were subsampled to contain 100,000 training sentences and 10,000 evaluation and testing sentences.
SNLI and Yelp have a respective vocabulary of 9,990 and 8,411 words and both have an average of 10 words per sentence.

\subsection{Evaluation metrics}
We use several metrics to evaluate the quality of the learned generative models.

\textbf{(Log-likelihood and perplexity per word)} The negative log-likelihood (NLL) $-\mathds{E}_{q_{\phi}(\vz | \vx)} p_{\theta}(\vx | \vz)$ indicates how well the model reconstructs the input sentence.
The perplexity per word (PPL) is the geometric mean of the reciprocal of the probability assigned to the correct word by the model.
Since we observe the opposite in the first case and the reciprocal in the second, we want to minimize both of these metrics.
They are approximated by sampling 100 times in the distribution $q$ for each sentence.

\textbf{(BLEU score)} For each sentence in the test set, a latent representation is sampled in the distribution $q$ and a sentence is generated without using teacher forcing from the representation. The BLEU score \cite{papieni2002} will represent the proportion of n-grams (for n going from 1 to 4) in the generated sentence that can be found in the original one. If the generated sentence is shorter than the original, a penalty is applied since it is easier to avoid mistakes when less words are produced.

\textbf{(Active units)} The number of active units (AU) represent the number of dimensions of the latent variable that co-vary with the observations. According to \citet{burda2016}, a greater number of active units is usually representative of a richer latent variable.
We follow their article and use a threshold value of $0.01$.
The dimension of the latent variable is 32, therefore we will have at most 32 active units.

\textbf{(Mutual Information)} We also report the mutual information between the latent variable and the output probability distribution.
A higher mutual information indicates that the latent variable is better used by the model.
We follow the methodology of \citet{he19}.

\subsection{Baseline}

We compare our approach to previous work including a "standard" VAE, the free bits technique \cite{kingma2016} and the pre-training approach \cite{li2019}:
\begin{itemize}
    \item \textbf{(Free bits)}
The free bits technique consists in introducing a constraint on the divergence term with the prior distribution so that it doesn't fall below a pre-determined threshold $\lambda$.
We follow previous work and fix $\lambda = 8$ in all of our experiments.
    \item \textbf{(Pre-training)}
This baseline consists in training the model as a classic autoencoder first. Then the decoder is reset and the model is trained as a VAE.
\end{itemize}

All of our experiments reweight the divergence term during the training of the model.
This technique, proposed by \citet{bowman2016}, consists in steadily increasing the reweighting factor from 0 to 1 during the first epochs of training.
The idea is to force the model to ignore the divergence with prior term in the beginning of training.

\subsection{Results and analysis}

\begin{table*}[t]
\centering
\scriptsize
~\\
\caption{
Impact of the fraternal dropout hyperparameter $\alpha$ over the Yelp dataset.
We want to maximize the metrics with a $\uparrow$ and minimize those with a $\downarrow$.}
~\\~\\
\begin{tabular}{c | c  c  c  c   c } 
 \hline
 \begin{tabular}{@{}c@{}}$\alpha$\end{tabular} & NLL $\downarrow$ & PPL $\downarrow$ & AU $\uparrow$ & MI $\uparrow$  & BLEU $\uparrow$ \\
 \hline
0.01 & \textbf{28.91} & \textbf{18.44} & 4 & 6.34 & 5.20 \\
0.1 & 29.50 & 19.12 & 5 & 7.03 & 6.40 \\
0.5 & 30.69 & 21.53 & \textbf{6} & \textbf{7.47} & \textbf{6.81} \\
1.0 & 32.30 & 25.28 & 4 & 6.87 & 6.03 \\
2.0 & 33.41 & 28.27 & 4 & 7.29 & 6.09 \\
\hline
\end{tabular}

\label{table:yelp_mse_1}
\end{table*}
\begin{table*}[t]
\centering
\scriptsize
~\\
\caption{%
Results of the various metrics over Yelp and SNLI for four different VAEs without and with fraternal dropout.
We want to maximize the metrics with a $\uparrow$ and minimize those with a $\downarrow$.
}
~\\~\\
\begin{tabular}{l|ccccc|ccccc} 
 & \multicolumn{5}{c|}{Yelp} & \multicolumn{5}{c}{SNLI}
 \\
 Configuration
 & NLL $\downarrow$ & PPL $\downarrow$ & UA $\uparrow$ & IM $\uparrow$ & BLEU $\uparrow$
 & NLL $\downarrow$ & PPL $\downarrow$ & UA $\uparrow$ & IM $\uparrow$ & BLEU $\uparrow$
 \\ \hline
Standard
& 33.40         & 28.25         & 2         & 1.14          & 1.43
& 32.57         & 20.64         & \bf{3}         & 0.52          & 2.32 
\\
+ fraternal dropout 
& \bf{29.50}    & \bf{19.12}    & \bf{5}    & \bf{7.03}     & \bf{6.40}
& \bf{30.01}    & \bf{16.28}    & 2         & \bf{4.75}     & \bf{5.73} 
\\ \hline
Free bits
& 29.54         & 19.20         & 32        & 5.69          & 4.02
& 28.88         & 14.66         & 32        & 4.63          & 4.77
\\
+ fraternal dropout 
& \bf{25.46}    & \bf{12.76}    & 32        & \bf{8.65}     & \bf{11.23}
& \bf{27.92}    & \bf{13.40}    & 32        & \bf{7.11}     & \bf{8.44} 
\\ \hline
Pre-train.
& 33.74         & 29.21         & 2         & 0.71          & 0.83
& 31.76         & 19.14         & 3         & 1.14          & 2.76 
\\
+ fraternal dropout 
& \bf{26.18}    & \bf{13.71}    & \bf{22}   & \bf{8.24}     & \bf{9.69}
& \bf{24.69}    & \bf{9.92}     & \bf{22}   & \bf{8.32}     & \bf{13.43} 
\\ \hline
Free bits + Pre-train.
& 25.93         & 13.37         & 32        & 8.14          & 7.54
& 23.33         & 8.75          & 32        & 8.49          & 13.9 
\\
+ fraternal dropout 
& \bf{23.63}    & \bf{10.62}    & 32        & \bf{8.81}     & \bf{13.54}
& \bf{21.00}    & \bf{7.04}     & 32        & \bf{9.07}     & \bf{21.35} 
\\ 
 \hline
\end{tabular}
\label{table:yelp}
\end{table*}

We first evaluate the impact of the fraternal dropout hyperparameter $\alpha$ on the Yelp dataset. Results are reported in Table~\ref{table:yelp_mse_1}.
We can observe that a compromise needs to be found between all the metrics,
i.e.\ we need to find a point of equilibrium between minimizing the NLL and the PPL and maximizing AU, MI and the BLEU score.
In the following experiments, we fixed $\alpha = 0.1$.

We report results in different configurations in Table \ref{table:yelp}.
Our approach yields improvements for all the metrics in all configurations for both datasets.
The posterior collapse problem is significant in configurations not using free bits or fraternal dropout, the mutual information being around 1 and the number of active units being 2.
Adding fraternal dropout results in a gain between 4 and 7 points for the mutual information in the two configurations.
In configurations where the model is pre-trained, we observe that this pre-training alone does not prevent the posterior collapse since there are only 2 and 3 active units respectively while the same model with fraternal dropout retains 22 active units.
Interestingly, our approach has a bigger impact than free bits on Yelp and a similar one on SNLI while using less active units in both cases. This is an indication that the free bits technique forces the latent variables to be decorrelated artificially, i.e.\ the decoder still ignores their values.

As explained previously, the BLEU score is computed over sentences generated without teacher forcing and therefore can also be used to estimate the quality of the latent representations.
Once again, fraternal dropout improves results on this metric.

\textbf{Interpolation}
We show some examples of sentences generated via interpolation between two latent representations sampled from the prior in Table~\ref{table:interpolation}.
We see that our method seems to produce coherent sentences with a gradual change in length and meaning between the successive sentences.

\begin{table*}[t]
\centering
\scriptsize
~\\
\caption{%
Examples of interpolations between two representations sampled from the prior for the configuration with a pre-training and the free bits over the SNLI dataset. The second example also includes fraternal dropout.
}
~\\~\\
\begin{tabular}{l|l} 
\hline
\bf{Free bits + Pre-train} &  \bf{Free bits + Pre-train + fraternal dropout} \\ \\
a boy is in front of a group of people.& the young boy is in a picture.\\
a man in a blue shirt is standing in front of a crowd of people. & the young child is in front of a mother.\\
a child in blue is holding a camera.& the small child is in front of a mother. \\
a child in blue pants holding a camera while another man watches. & a small child in pink holds a picture of her mother. \\
a child in blue pants holding a camera while another man in a black shirt looks on. & a small child in pink sits in a picture with her mother. \\
 \hline
\end{tabular}
\label{table:interpolation}
\end{table*}

\section{Conclusion}

In this work, we propose to rely on parameter regularization to prevent the posterior collapse problem in VAEs.
This approach is different from previous work in the literature.
We observe that our approach has two benefits: it improves the quality of generated text and increases the use of the latent variable.
Future works could explore other methods for parameter regularization \cite{arpit2019hdetach, krueger2017zoneout, gal2016theoretically}.

\begin{ack}
We thank François Yvon and Matthieu Labeau for proofreading the article.
This work benefited from computations done on the Saclay-IA platform and an access to the computational resources of IDRIS through the resource allocation 20XX-AD11011600 attributed by GENCI.
\end{ack}

\bibliography{biblio}

\begin{thebibliography}{23}
\providecommand{\natexlab}[1]{#1}
\providecommand{\url}[1]{\texttt{#1}}
\expandafter\ifx\csname urlstyle\endcsname\relax
  \providecommand{\doi}[1]{doi: #1}\else
  \providecommand{\doi}{doi: \begingroup \urlstyle{rm}\Url}\fi

\bibitem[Bowman et~al.(2015)Bowman, Angeli, Potts, and Manning]{snli}
S.~R. Bowman, G.~Angeli, C.~Potts, and C.~D. Manning.
\newblock A large annotated corpus for learning natural language inference.
\newblock In \emph{Proceedings of the 2015 Conference on Empirical Methods in
  Natural Language Processing (EMNLP)}. Association for Computational
  Linguistics, 2015.

\bibitem[Bowman et~al.(2016)Bowman, Vilnis, Vinyals, Dai, Jozefowicz, and
  Bengio]{bowman2016}
S.~R. Bowman, L.~Vilnis, O.~Vinyals, A.~Dai, R.~Jozefowicz, and S.~Bengio.
\newblock Generating sentences from a continuous space.
\newblock In \emph{Proceedings of The 20th {SIGNLL} Conference on Computational
  Natural Language Learning}, pages 10--21, Berlin, Germany, Aug. 2016.
  Association for Computational Linguistics.
\newblock \doi{10.18653/v1/K16-1002}.
\newblock URL \url{https://www.aclweb.org/anthology/K16-1002}.

\bibitem[Burda et~al.(2016)Burda, Grosse, and Salakhutdinov]{burda2016}
Y.~Burda, R.~B. Grosse, and R.~Salakhutdinov.
\newblock Importance weighted autoencoders.
\newblock In Y.~Bengio and Y.~LeCun, editors, \emph{Proceedings of 4th
  International Conference on Learning Representations}, 2016.
\newblock URL \url{http://arxiv.org/abs/1509.00519}.

\bibitem[Dempster et~al.(1977)Dempster, Laird, and Rubin]{dempster1977em}
A.~P. Dempster, N.~M. Laird, and D.~B. Rubin.
\newblock Maximum likelihood from incomplete data via the em algorithm.
\newblock \emph{Journal of the Royal Statistical Society: Series B
  (Methodological)}, 39\penalty0 (1):\penalty0 1--22, 1977.

\bibitem[Dieng et~al.(2019)Dieng, Kim, Rush, and Blei]{dieng2019skipvae}
A.~B. Dieng, Y.~Kim, A.~M. Rush, and D.~M. Blei.
\newblock Avoiding latent variable collapse with generative skip models.
\newblock In K.~Chaudhuri and M.~Sugiyama, editors, \emph{Proceedings of
  Machine Learning Research}, volume~89 of \emph{Proceedings of Machine
  Learning Research}, pages 2397--2405. PMLR, 16--18 Apr 2019.
\newblock URL \url{http://proceedings.mlr.press/v89/dieng19a.html}.

\bibitem[Dozat and Manning(2017)]{dozat2017biaffine}
T.~Dozat and C.~D. Manning.
\newblock Deep biaffine attention for neural dependency parsing.
\newblock In \emph{5th International Conference on Learning Representations,
  {ICLR} 2017, Toulon, France, April 24-26, 2017, Conference Track
  Proceedings}. OpenReview.net, 2017.
\newblock URL \url{https://openreview.net/forum?id=Hk95PK9le}.

\bibitem[Gal and Ghahramani(2016)]{gal2016theoretically}
Y.~Gal and Z.~Ghahramani.
\newblock A theoretically grounded application of dropout in recurrent neural
  networks.
\newblock In D.~Lee, M.~Sugiyama, U.~Luxburg, I.~Guyon, and R.~Garnett,
  editors, \emph{Advances in Neural Information Processing Systems}, volume~29,
  pages 1019--1027. Curran Associates, Inc., 2016.
\newblock URL
  \url{https://proceedings.neurips.cc/paper/2016/file/076a0c97d09cf1a0ec3e19c7f2529f2b-Paper.pdf}.

\bibitem[Goodfellow et~al.(2014)Goodfellow, Pouget{-}Abadie, Mirza, Xu,
  Warde{-}Farley, Ozair, Courville, and Bengio]{goodfellow2014gan}
I.~J. Goodfellow, J.~Pouget{-}Abadie, M.~Mirza, B.~Xu, D.~Warde{-}Farley,
  S.~Ozair, A.~C. Courville, and Y.~Bengio.
\newblock Generative adversarial nets.
\newblock In Z.~Ghahramani, M.~Welling, C.~Cortes, N.~D. Lawrence, and K.~Q.
  Weinberger, editors, \emph{Advances in Neural Information Processing Systems
  27: Annual Conference on Neural Information Processing Systems 2014, December
  8-13 2014, Montreal, Quebec, Canada}, pages 2672--2680, 2014.
\newblock URL
  \url{http://papers.nips.cc/paper/5423-generative-adversarial-nets}.

\bibitem[Havrylov and Titov(2020)]{havrylov2020vae}
S.~Havrylov and I.~Titov.
\newblock Preventing posterior collapse with levenshtein variational
  autoencoder.
\newblock \emph{arXiv preprint arXiv:2004.14758}, 2020.

\bibitem[He et~al.(2019)He, Spokoyny, Neubig, and Berg{-}Kirkpatrick]{he19}
J.~He, D.~Spokoyny, G.~Neubig, and T.~Berg{-}Kirkpatrick.
\newblock Lagging inference networks and posterior collapse in variational
  autoencoders.
\newblock In \emph{Proceedings of the 7th International Conference on Learning
  Representations}. OpenReview.net, 2019.
\newblock URL \url{https://openreview.net/forum?id=rylDfnCqF7}.

\bibitem[Kanuparthi et~al.(2019)Kanuparthi, Arpit, Kerg, Ke, Mitliagkas, and
  Bengio]{arpit2019hdetach}
B.~Kanuparthi, D.~Arpit, G.~Kerg, N.~R. Ke, I.~Mitliagkas, and Y.~Bengio.
\newblock h-detach: Modifying the {LSTM} gradient towards better optimization.
\newblock In \emph{7th International Conference on Learning Representations,
  {ICLR} 2019, New Orleans, LA, USA, May 6-9, 2019}. OpenReview.net, 2019.
\newblock URL \url{https://openreview.net/forum?id=ryf7ioRqFX}.

\bibitem[Kingma and Welling(2014)]{kingma2013}
D.~P. Kingma and M.~Welling.
\newblock Auto-encoding variational bayes.
\newblock In Y.~Bengio and Y.~LeCun, editors, \emph{2nd International
  Conference on Learning Representations, {ICLR} 2014, Banff, AB, Canada, April
  14-16, 2014, Conference Track Proceedings}, 2014.
\newblock URL \url{http://arxiv.org/abs/1312.6114}.

\bibitem[Kingma et~al.(2016)Kingma, Salimans, Jozefowicz, Chen, Sutskever, and
  Welling]{kingma2016}
D.~P. Kingma, T.~Salimans, R.~Jozefowicz, X.~Chen, I.~Sutskever, and
  M.~Welling.
\newblock Improved variational inference with inverse autoregressive flow.
\newblock In D.~D. Lee, M.~Sugiyama, U.~V. Luxburg, I.~Guyon, and R.~Garnett,
  editors, \emph{Advances in Neural Information Processing Systems 29}, pages
  4743--4751. Curran Associates, Inc., 2016.
\newblock URL
  \url{http://papers.nips.cc/paper/6581-improved-variational-inference-with-inverse-autoregressive-flow.pdf}.

\bibitem[Krueger et~al.(2016)Krueger, Maharaj, Kram{\'{a}}r, Pezeshki, Ballas,
  Ke, Goyal, Bengio, Larochelle, Courville, and Pal]{krueger2017zoneout}
D.~Krueger, T.~Maharaj, J.~Kram{\'{a}}r, M.~Pezeshki, N.~Ballas, N.~R. Ke,
  A.~Goyal, Y.~Bengio, H.~Larochelle, A.~C. Courville, and C.~Pal.
\newblock Zoneout: Regularizing rnns by randomly preserving hidden activations.
\newblock \emph{CoRR}, abs/1606.01305, 2016.
\newblock URL \url{http://arxiv.org/abs/1606.01305}.

\bibitem[Li et~al.(2019)Li, He, Neubig, Berg-Kirkpatrick, and Yang]{li2019}
B.~Li, J.~He, G.~Neubig, T.~Berg-Kirkpatrick, and Y.~Yang.
\newblock A surprisingly effective fix for deep latent variable modeling of
  text.
\newblock In \emph{Proceedings of the 2019 Conference on Empirical Methods in
  Natural Language Processing and the 9th International Joint Conference on
  Natural Language Processing (EMNLP-IJCNLP)}, pages 3603--3614, Hong Kong,
  China, Nov. 2019. Association for Computational Linguistics.
\newblock \doi{10.18653/v1/D19-1370}.
\newblock URL \url{https://www.aclweb.org/anthology/D19-1370}.

\bibitem[Livne et~al.(2020)Livne, Swersky, and Fleet]{livne2020sentencemim}
M.~Livne, K.~Swersky, and D.~J. Fleet.
\newblock Sentencemim: A latent variable language model.
\newblock \emph{arXiv preprint arXiv:2003.02645}, 2020.

\bibitem[Merity et~al.(2018)Merity, Keskar, and Socher]{merity2017lm}
S.~Merity, N.~S. Keskar, and R.~Socher.
\newblock Regularizing and optimizing lstm language models.
\newblock In \emph{Proceedings of ICLR 2018}, 2018.

\bibitem[Papineni et~al.(2002)Papineni, Roukos, Ward, and Zhu]{papieni2002}
K.~Papineni, S.~Roukos, T.~Ward, and W.-J. Zhu.
\newblock {B}leu: a method for automatic evaluation of machine translation.
\newblock In \emph{Proceedings of the 40th Annual Meeting of the Association
  for Computational Linguistics}, pages 311--318. Association for Computational
  Linguistics, July 2002.
\newblock \doi{10.3115/1073083.1073135}.
\newblock URL \url{https://www.aclweb.org/anthology/P02-1040}.

\bibitem[Pelsmaeker and Aziz(2020)]{pelsmaeker2019vae}
T.~Pelsmaeker and W.~Aziz.
\newblock Effective estimation of deep generative language models.
\newblock In \emph{Proceedings of the 58th Annual Meeting of the Association
  for Computational Linguistics}, pages 7220--7236, Online, July 2020.
  Association for Computational Linguistics.
\newblock \doi{10.18653/v1/2020.acl-main.646}.
\newblock URL \url{https://www.aclweb.org/anthology/2020.acl-main.646}.

\bibitem[Shen et~al.(2017)Shen, Lei, Barzilay, and Jaakkola]{shen2017}
T.~Shen, T.~Lei, R.~Barzilay, and T.~Jaakkola.
\newblock Style transfer from non-parallel text by cross-alignment.
\newblock In I.~Guyon, U.~V. Luxburg, S.~Bengio, H.~Wallach, R.~Fergus,
  S.~Vishwanathan, and R.~Garnett, editors, \emph{Advances in Neural
  Information Processing Systems}, volume~30, pages 6830--6841. Curran
  Associates, Inc., 2017.
\newblock URL
  \url{https://proceedings.neurips.cc/paper/2017/file/2d2c8394e31101a261abf1784302bf75-Paper.pdf}.

\bibitem[Vaswani et~al.(2017)Vaswani, Shazeer, Parmar, Uszkoreit, Jones, Gomez,
  Kaiser, and Polosukhin]{vaswani2017transformers}
A.~Vaswani, N.~Shazeer, N.~Parmar, J.~Uszkoreit, L.~Jones, A.~N. Gomez, L.~u.
  Kaiser, and I.~Polosukhin.
\newblock Attention is all you need.
\newblock In I.~Guyon, U.~V. Luxburg, S.~Bengio, H.~Wallach, R.~Fergus,
  S.~Vishwanathan, and R.~Garnett, editors, \emph{Advances in Neural
  Information Processing Systems}, volume~30. Curran Associates, Inc., 2017.
\newblock URL
  \url{https://proceedings.neurips.cc/paper/2017/file/3f5ee243547dee91fbd053c1c4a845aa-Paper.pdf}.

\bibitem[Yang et~al.(2017)Yang, Hu, Salakhutdinov, and
  Berg-Kirkpatrick]{yang2017cnnvae}
Z.~Yang, Z.~Hu, R.~Salakhutdinov, and T.~Berg-Kirkpatrick.
\newblock Improved variational autoencoders for text modeling using dilated
  convolutions.
\newblock In D.~Precup and Y.~W. Teh, editors, \emph{Proceedings of the 34th
  International Conference on Machine Learning}, volume~70 of \emph{Proceedings
  of Machine Learning Research}, pages 3881--3890, International Convention
  Centre, Sydney, Australia, 06--11 Aug 2017. PMLR.
\newblock URL \url{http://proceedings.mlr.press/v70/yang17d.html}.

\bibitem[Zolna et~al.(2018)Zolna, Arpit, Suhubdy, and
  Bengio]{zolna2018fraternal}
K.~Zolna, D.~Arpit, D.~Suhubdy, and Y.~Bengio.
\newblock Fraternal dropout.
\newblock In \emph{International Conference on Learning Representations}, 2018.
\newblock URL \url{https://openreview.net/forum?id=SJyVzQ-C-}.

\end{thebibliography}

\end{document}